\documentclass[lettersize,journal]{IEEEtran}
\usepackage{cite}
\usepackage{amsmath,amssymb,amsfonts}
\usepackage{algorithmic}
\usepackage{graphicx}
\usepackage{textcomp}

\usepackage{amsmath,amsfonts}
\usepackage{algorithmic}
\usepackage{array}

\usepackage{stfloats}
\usepackage{url}
\usepackage{verbatim}
\usepackage{graphicx}
\usepackage{cite}

\usepackage{soul, color, xcolor}

\usepackage{graphics}
\usepackage{epsfig}
\usepackage{mathptmx} 
\usepackage{times} 
\usepackage{amssymb}
\usepackage{float}  
\usepackage{amsmath,bm}
\usepackage{cite}
\usepackage{CJKutf8}
\usepackage{booktabs}
\usepackage{url}
\usepackage{bm}

\usepackage{multirow}
\usepackage{booktabs}
\usepackage{epstopdf}
\usepackage{epsfig}
\usepackage{longtable}
\usepackage{supertabular}

\usepackage{changepage}
\usepackage{verbatim}
\usepackage{xcolor}
\usepackage{subfig}
\usepackage{float}
\usepackage{stfloats}
\usepackage[mathscr]{euscript}

\usepackage{array}
\usepackage{url}
\usepackage{tikz,xcolor}
\usepackage[ruled,vlined]{algorithm2e}
\usepackage{times}
\usepackage{cancel}
\usepackage{balance}
\usepackage{color}
\usepackage{mathtools}
\usepackage[ruled,vlined]{algorithm2e}
\usepackage{bm}

\usepackage{diagbox}
\usepackage{float}
\usepackage{epstopdf}
\usepackage{pifont}
\usepackage{fixltx2e}
\usepackage{multirow}
\usepackage{booktabs}
\usepackage{url}
\usepackage{svg}
\usepackage{threeparttable}
\usepackage[linkcolor=black,citecolor=black,urlcolor=black,colorlinks=true]{hyperref}
\usepackage{subcaption}
\usepackage{slashbox}
\usepackage{makecell}
\usepackage[utf8]{inputenc}
\usepackage{textgreek}

\begin{document}

\title{WD-DETR: Wavelet Denoising-Enhanced Real-Time Object Detection Transformer for Robot Perception with Event Cameras}

\author{Yangjie Cui, Boyang Gao, Yiwei Zhang, Xin Dong, Jinwu Xiang, Daochun Li, Zhan Tu$^*$
        
\thanks{Yangjie Cui, Yiwei Zhang, and Daochun Li are with the School of Aeronautic Science and Engineering,
Beihang University, Bejing 100191, China (e-mail: cuiyangjie@buaa.edu.cn, zhangyiwei@buaa.edu.cn, lidc@buaa.edu.cn).}
\thanks{Boyang Gao and Xin Dong are the Hangzhou Innovation Institute of Beihang University, Yuhang District, Hangzhou 310023, China (e-mail: bygao@buaa.edu.cn, xindong324@buaa.edu.cn).}
\thanks{Jinwu Xiang and Zhan Tu are the Institute of Unmanned System, Beihang University, Beijing 100191, China. Jingwu Xiang is also with the Tianmushan Laboratory Xixi Octagon City, Yuhang District, Hangzhou 310023, China (e-mail: xiangjw@buaa.edu.cn, zhantu@buaa.edu.cn).}
\thanks{$^*$ To whom the correspondence should be addressed.}
}
\markboth{Journal of \LaTeX\ Class Files,~Vol.~14, No.~8, December~2024}%
{Shell \MakeLowercase{\textit{et al.}}: A Sample Article Using IEEEtran.cls for IEEE Journals}
\maketitle
\begin{abstract}
Leveraging low latency and high dynamic range advantages, event-based vision shows superior effectiveness for robot perception tasks, especially in scenarios involving rapid object motion, low-light conditions, or sudden changes in illumination. 
Event cameras operate by detecting brightness changes asynchronously at each pixel, resulting in a sparse and noisy output.
Previous studies on event camera sensing have demonstrated certain detection performance using dense event representations. However, the accumulated noise in such dense representations has received insufficient attention, which degrades the representation quality and increases the likelihood of missed detections.
To address this challenge, we propose the Wavelet Denoising-enhanced DEtection TRansformer, i.e., WD-DETR network, for event cameras. In particular, a dense event representation is presented firstly that enables real-time reconstruction of events as tensors. Then, a wavelet transform method is designed to filter noise in the event representations. Such a method is integrated into the backbone for feature extraction. The extracted features are subsequently fed into a transformer-based network for object prediction. To further reduce inference time, we incorporate the Dynamic Reorganization Convolution  Block (DRCB) as a fusion module within the hybrid encoder.
The proposed method has been evaluated on three event-based object detection datasets—DSEC, Gen1, and 1Mpx—and the results demonstrate that WD-DETR outperforms tested state-of-the-art methods. Additionally, We implement our approach on a common onboard computer for robots, the NVIDIA Jetson Orin NX, achieving a high frame rate of approximately 35 FPS using TensorRT FP16, which is exceptionally well-suited for real-time perception of onboard robotic systems. In outdoor flight experiments conducted under low light conditions of approximately 10 Lux, WD-DETR demonstrated effective real-time object detection capabilities.
\end{abstract}


\section{Introduction}
\label{sec:introduction}
Object detection is widely applied across various fields, with RGB cameras being the most commonly employed in robotic systems. However, RGB cameras face significant challenges in scenarios involving rapid motion, low-light conditions, or abrupt changes in illumination. Performance degradation primarily arises from the need for adequate exposure time and the large volume of synchronous output data. In high-speed movements, motion blur compromises image clarity, making accurate object detection difficult. Similarly, extreme lighting conditions, such as high-contrast environments, can cause overexposure or underexposure, further limiting the effectiveness of RGB cameras. As a result, alternative sensor modalities, such as event cameras, are essential for ensuring reliable performance under these challenging conditions.

Event cameras operate by asynchronously detecting changes in pixel intensity, capturing dynamic information exclusively rather than traditional frame-based images. Each sensor pixel independently responds to variations in brightness, enabling these cameras to achieve high temporal resolution, low latency, and reduced data redundancy. These characteristics enable it particularly effective in scenarios involving high-speed motion or special light environments \cite{Event-Nature}, where conventional RGB cameras encounter challenges. The event camera responds to brightness changes independently and asynchronously to each pixel in the scene, inherently resulting in sparse and noise-rich output \cite{E-MLB_noise}. 

\begin{figure}[tbp!]
    \centering
    \includegraphics[width=1.0\linewidth]{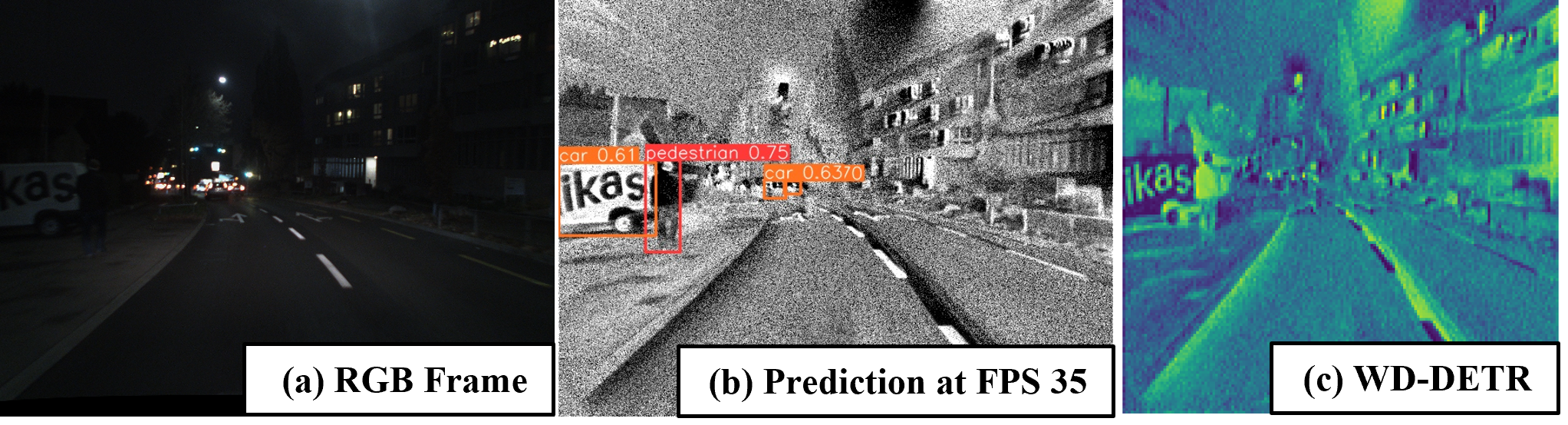}
    \caption{Visualization of the dense event representation and detection results on the DSEC dataset. (a) is the RGB frame, in which the person can not be seen. (b) is the dense event representation and prediction results by WD-DETR at FPS 35, in which the person is evident. (c) represents the dense event features extracted by WD-DETR in stage 0, where preliminary wavelet filtering is applied to effectively reduce noise.}
    \label{fig:vis}
    \vspace{-0.5cm}
\end{figure}

For the sparse event data, there are two main categories of methods: those that utilize the sparsity directly and those that convert it into a dense representation. Sparse representation retains the original characteristics of the event information, and the amount of data is small. The direct processing of sparse event streams is mainly based on asynchronous spiking neural network (SNN) \cite{SNN_learning_2020, SNN_backpro_2020}, and Graph Neural Network (CNN) \cite{GNN_2020, GNN_2021, GNN_aegnn_2022, GNN_voxel_2022}. SNN utilizes the asynchrony of the event camera for object detection to improve efficiency and performance. Due to the lack of corresponding hardware support for SNN at present, the performance and application are limited. GNN significantly reduces the computation and latency of event-by-event processing by recalculating network activations on all nodes affected by new events. Due to the sparse event stream, it is still less accurate than dense methods for event-based vision.

In contrast, dense representation methods offer improved performance since they can leverage mature machine learning algorithms and neural network architectures. This kind of method first segments the event stream into groups according to time and then densely represents the event polarity as tensors, which supports network operations such as convolution. 
\cite{GET-Net, SSM_2024, GWD-Net, RecurrentTransformerNet} combine different event representations with mature neural network architectures and achieve competitive results. However, very few studies have focused on the noise in dense event representations (Fig. \ref{fig:vis}). The reason why there is noise in the event stream is that individual pixels in the event camera respond independently and asynchronously to changes in light intensity, which causes some pixels to be recorded due to false triggers but not corrected \cite{E-MLB_noise}. After a dense representation, the event information accumulates and becomes rich, but the noise also accumulates at the same time. Therefore, it is essential to filter the dense event representation, as the accrued noise will affect the accuracy of the feature extraction results in the subsequent network. 

To address the challenge of noise interference in dense event representations for object detection, we propose the Wavelet Denoising-enhanced DEtection TRansformer, i.e., WD-DETR network, for onboard event cameras. First, event data is densely represented as grayscale images using a lightweight, high-fidelity intensity reconstruction scheme. The wavelet transform is then integrated into the backbone network to filter noise, leveraging its strong time-frequency analysis capabilities, which have proven effective in signal processing and pattern recognition. 
To further reduce inference time, the RepConv model is incorporated into the backbone, while the Dynamic Reorganization Conv Block (DRCB) is employed in the fusion block of the hybrid encoder module. 
Our method is evaluated on three event-based object detection datasets: DSEC \cite{DSEC-Dataset}, Gen1 \cite{Gen1-Dataset}, and 1Mpx \cite{1mpx_dataset}. The proposed WD-DETR achieves mean Average Precision (mAP) scores of $38.4\%$, $52.8\%$ and $65.5\%$, respectively, surpassing other state-of-the-art methods involved in the eveluation. Additionally, WD-DETR achieves a real-time processing speed of approximately 35 frames per second on the onboard computing platform NVIDIA Jetson Orin NX. The WD-DETR method can effectively detect the objects in the outdoor flight experiments at a low light intensity of 10 Lux.

 The remainder of this paper is organized as follows. Section \ref{sec:related_works} addresses the related work of object detection methods for event cameras and transformer networks in RGB image detection. Section \ref{sec:method} presents the framework of the proposed method. Section \ref{sec:experiments} presents and discusses the experiments results. Finally, Section \ref{sec:conclusions} summarizes our work.

\section{RELATED WORK}
\label{sec:related_works}

In this section, the related works have been introduced respectively, which include the classical learning methods for object detection of event camera, the application of transformers in computer vision, and the wavelet transform method in deep learning.

\subsection{Object Detection Method for Event Camera}
\label{subsec:ref_ob}
The object detection methods based on event cameras can be divided into events-frames fusion and only events. Events-frames fusion methods \cite{Event-Nature, Retina-Net_2022, Event_RGB_2023, CAFR_2024} can achieve higher accuracy, but the fusion method of events and frames is more complicated as the network needs to process the events and frames simultaneously. 
The methods of object detection that rely only on event camera can be divided into sparse event representation method and dense event representation method. For the sparse event representation, graph neural network (GNN)  \cite{GNN_pushing_2022, GNN_memory_2023, GNN_event_2023} and spike neural network (SNN) \cite{SNN_learning_2020, SNN_backpro_2020} are commonly used to extract event features. 
As both of these methods lack the corresponding hardware acceleration support, the inference time are limited by the airborne computing resources. 
The object detection method based on the dense representation \cite{
GET-Net, SSM_2024, GWD-Net, RecurrentTransformerNet, DE_sodformer_2023}  of events first constructs the event flow into a dense tensor (event representation) according to the time window, which supports basic operations of neural networks such as convolution. 
We follow this line of works, but pay more attention to the cumulative noise generated in the dense event representation in order to build a fast, lightweight, and high-performance specification framework for object detection.

\subsection{Transformer Network}
\label{subsec:ref_tm}
The core strength of the Transformer network is its attention mechanism \cite{Attention}, which efficiently allows the model to selectively focus on the most important parts of the current context when working with sequences. 

Vision Transformer \cite{VIT} is a pure transformer directly applied to the sequence of image patches for image classification tasks. Based on this work, the detection transformer (DETR) \cite{DETR} redesigns the framework of object detection, a simple and fully end-to-end object detector, treats the object detection task as an intuitive set prediction problem, eliminating traditional hand-crafted components such as anchor generation and non-maximum suppression (NMS) post-processing. Then, according to the long training schedule and poor performance for small objects, some improvements are taken in DETR architecture, such as Deformable DETR \cite{Deformable_DETR}. 

\subsection{Wavelet in deep learning}
\label{subsec:ref_wave}
As the event representation or images can be treated as signals, signal processing techniques, such as wavelet transform \cite{WT_denosing_2011, wavelet_tip_1, wavelet_tip_2}, are effective in low-level tasks for reducing noise while preserving important image details. These methods \cite{WT-CNN, WT-diffusion, WT_edge_2023, WT_hyperspectral_2022, WaveCNet_2021} leverage the multi-resolution analysis capabilities of wavelets, which allow for the decomposition of an image into different frequency components. Wavelet transform method enables targeted noise reduction, particularly in the high-frequency components where noise is often concentrated, without significantly affecting the low-frequency components that carry the essential image structure.

In current deep learning, while wavelet transformer is commonly applied as image preprocessing or post-processing \cite{wavelet_1_2017, wavelet_2_2018, wavelet_3_2019}, it is also used as down-sampling or up-sampling operations to design deep networks \cite{wavelet_4_2018, wavelet_5_2018, wavelet_6_2017, wavelet_7_2019, EWT_2024}. 
Inspired these studies, we combine wavelet transform into downsampling to improve the de-noising effect of event representation in the backbone network.

\section{WD-DETR NETWORK}
\label{sec:method}

In this section, the main architecture of WD-DETR for real-time object detection of event camera is illustrated firstly in section \ref{subsec:main_archi}. Next, the event representation method (Sec. \ref{subsec:event_repre}), WP-RepConv backbone (Sec. \ref{subsec:backbone}), hybrid encoder (Sec. \ref{subsec:encoder}) and matching part (Sec. \ref{subsec:decoder}) of the architecture are presneted in detail. 
\begin{figure*}[tbp!]
    \centering
    \includegraphics[width=.9\linewidth]{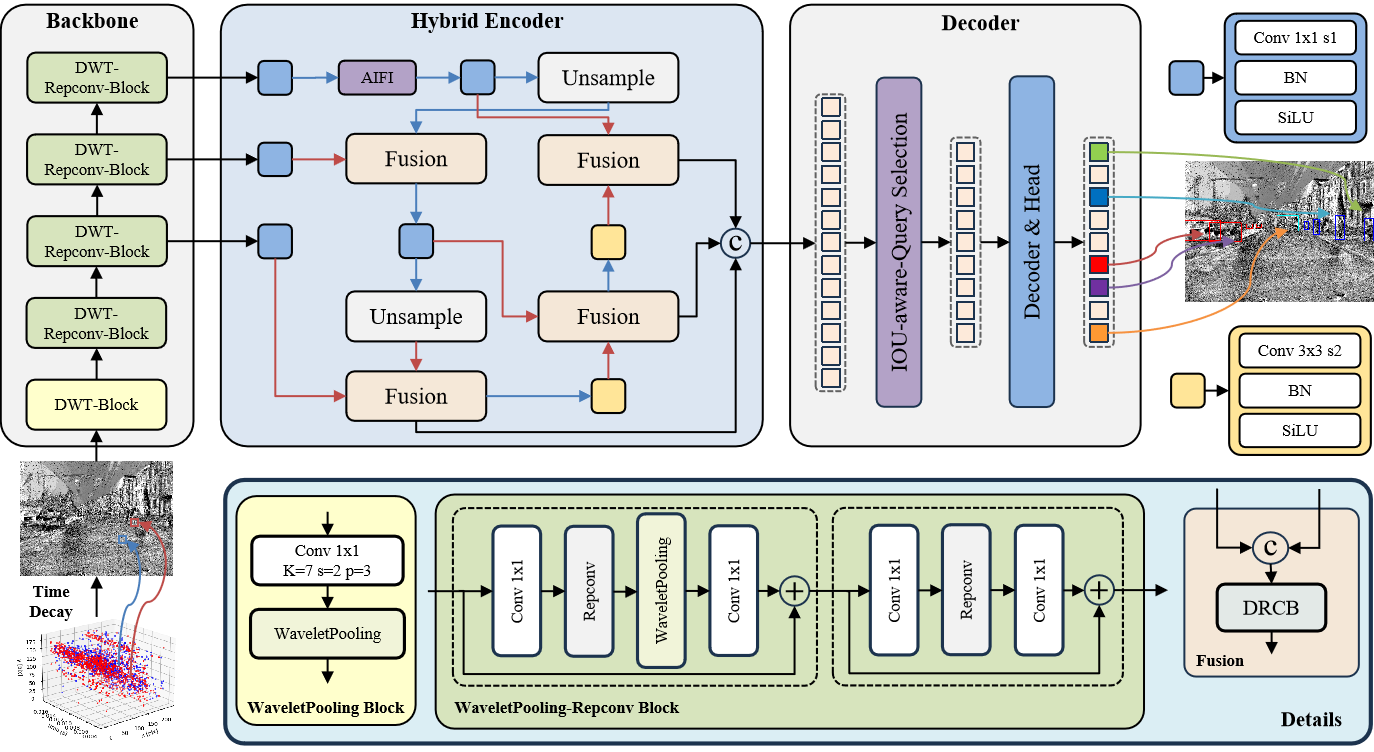}
    \caption{Framework of proposed WD-DETR. The event stream is split into several groups, and then be represented as a tensor through time decay accumulate method. Then it is fed into the backbone network for feature extraction which is composed by WP-Block and WP-RepConv Block. The last three stage features of backbone are then transfered into the hybrid encoder and matching part for object detection, specifying the bounding box and classification. }
    \label{fig:framework}
    \vspace{-0.5cm}
\end{figure*}

\subsection{Main Architecture}
\label{subsec:main_archi}
The proposed WD-DETR is a transformer-based network, the architecture of which is illustrated in Section \ref{fig:framework}. Four modules are utilized within the WD-DETR: an event representation, a backbone, a hybrid encoder, and a matching part. Specifically, the event stream is divided into groups with time windows, then a novel event representation method is proposed to meet the airborne memory requirements and avoid noise accumulation. Then, an efficient backbone enhanced by the wavelet transform and reprameterized convolution is proposed, which could extract the feature of event representation while denoising. Thirdly, the backbone's last three stage features are fed into the hybrid encoder, which could transform multiscale features into a sequence of features through the interscale feature interaction module and fusion module. We redesign the fusion module in a hybrid encoder using the Dynamic Reorganization Conv Block (DRCB) to reduce computational burden and improve accuracy. Subsequently, in matching part, the uncertainty-minimal query selection is employed to select a fixed number of encoder features to serve as initial object queries for the decoder. Finally, the decoder with auxiliary prediction heads iteratively optimizes object queries to generate categories and boxes.

\subsection{Event Representation}
\label{subsec:event_repre}

When the change in logarithm of light intensity measured by one of the pixel values of the event camera reaches a certain threshold $C$, the event camera outputs an event, denoted as 
\begin{align}
    e = [x, y, p, t], 
\end{align}
where $x$ and $y$ represent the pixel coordinates of the trigger event, $t$ is the timestamp of the trigger event, $p = {-1, +1}$ represents the polarity of the event, namely $+1$ represents an increase in brightness, and $+1$ represents a decrease in brightness.

The sensitivity of event cameras to changes in light intensity results in particularly severe noise \cite{E-MLB_noise}. To reduce the noise in dense event representation, the time decay accumulation method is proposed for event representation. As time passes, the stream of historical events has less impact on the current representation, so the cumulative weight decreases over time. Here, the decay function $d(t)$ is defined as an exponentially weighted polynomial function, formulated as:
\begin{align}
    d(t) = (1-k \time \Delta t)^b, \,\ k, b >0, 
\end{align}
where $k$ and $b$ are the parameters for decay. $\Delta t$ is the time interval for event data processing.

The intensity of each pixel can be re-formulated as:
\begin{align}
    S(x, y, t) = (S(x, y, t-\Delta t) + P_t \times C) \times d(t),
\end{align}
where $P_t$ is the event polarty matrix at time $t$.
Then, the event representation is limited to a certain range to increase robustness to drastic changes in environment illumination. 
\begin{align}
    S(x, y, t) = \text{clip} (S(x, y, t),\,\ S_{min }, \,\  S_{max}),
\end{align}
where $S_{max}$ and $S_{min}$ are the upper and lower bounds.
Finally, it is converted into an 8-bit grayscale image using linear mapping method, which is formulated as:
\begin{align}
    B(x, y, t) = 255 \times {(S(x, y, t) - S_{min}) / (S_{max} - S_{min})}
\end{align}

\subsection{Wavelet-enhanced Backbone}
\label{subsec:backbone}
\begin{figure}[!t]
    \centering
    \includegraphics[width=.9\linewidth]{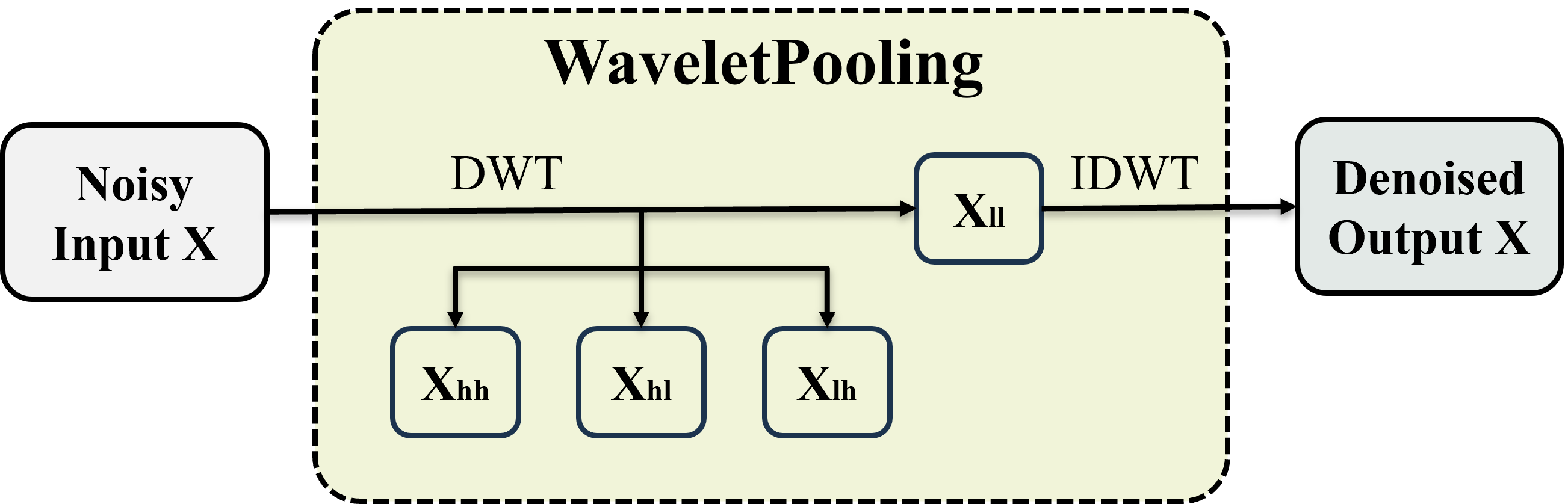}
    \caption{The approach of wavelet pooling. By applying the discrete wavelet transform to a noisy input $X$, the high-frequency components $X_{lh}$, $X_{hl}$, $X_{hh}$ are removed to suppress noise, and the inverse discrete wavelet transform is subsequently performed exclusively on the low-frequency components $X_{ll}$, thereby reducing the data volume by half.}
    \label{fig:waveletpooling}
    \vspace{-0.5cm}
\end{figure}

\begin{figure*}[h]
    \centering
    \includegraphics[width=.9\linewidth]{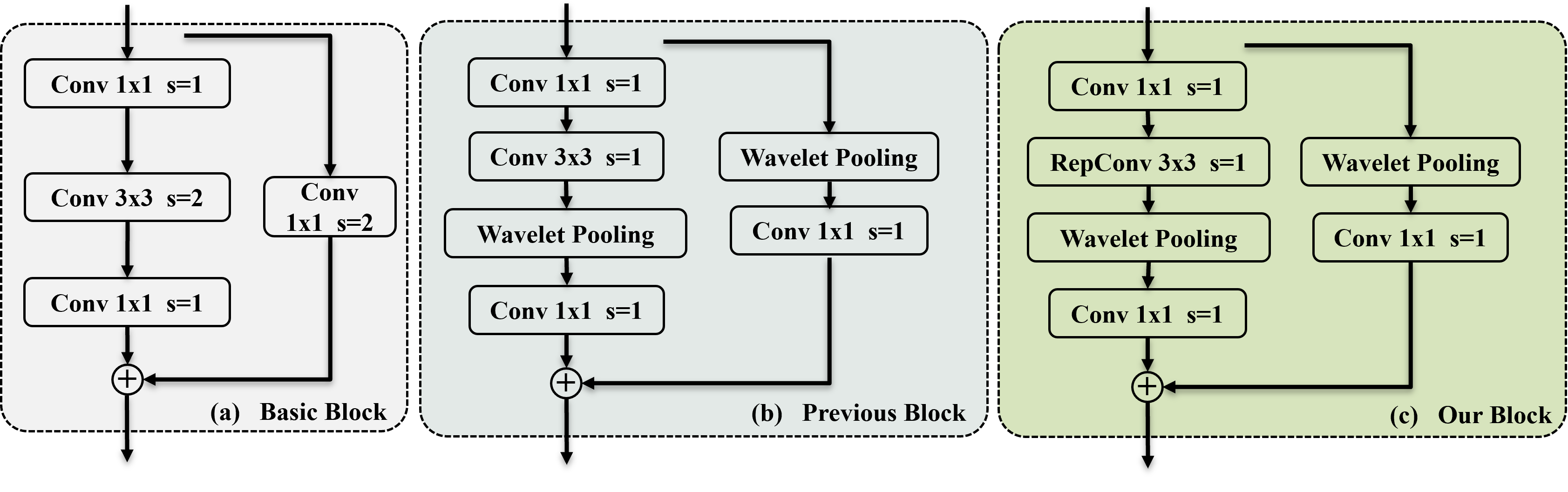}
    \caption{Diverse configurations of residual modules. (a) is the basic residual block, (b) is the previous denoising block, (c) is our denoising block.}
    \label{fig:WT-Repconv}
\end{figure*}

Dense event representations can extract richer features than sparse methods, leading to superior performance in object detection.
However, the process of obtaining dense event representations can lead to noise accumulation, which may degrade the quality of features extracted by the backbone network. Inspired by \cite{li2021wavecnet}, we propose the discrete wavelet transform (DWT) to reduce the adverse effect of noise on the features of events, as events could be seen as signals. Due to the increase in parameters caused by the WP block, we apply Reparameterized Convolution (RepConv) block \cite{ding2021repvgg} to replace some of the regular convolution block, which can reduce the inference time of the proposed network. In general, by combining wavelet transform downsampling with RepConv Block, the influence of noise on feature extraction is reduced and the parameters of network are not increased. 

The core of our network is wavelet pooling. Wavelet Pooling relies on the wavelet transform theory and is implemented using the Discrete Wavelet Transform (DWT) and the Inverse Discrete Wavelet Transform (IDWT). 

\textbf{1D DWT/IDWT.} Through one-dimensional discrete wavelet transform (DWT), it can decompose a given one-dimensional signal 
\[ x = \{ x_j \}_{j \in \mathbb{Z}} \]
into its low-frequency part 
\[ x_{\text{low}} = \{ x_k^{(\text{low})} \}_{k \in \mathbb{Z}} \]
and high-frequency part 
\[ x_{\text{high}} = \{ x_k^{(\text{high})} \}_{k \in \mathbb{Z}}. \]
The terms in $x_{\text{low}}$ and $x_{\text{high}}$ are respectively associated with the low-pass filter and high-pass filter of the orthogonal wavelet, so the DWT consists of filtering and downsampling. And IDWT reconstructs the vector \(x\) using $x_{\text{low}}$ and $x_{\text{high}}$, as follows:

\begin{equation}
x_{\text{low}} = \mathcal{L} x, \quad x_{\text{high}} = \mathcal{H} x,
\end{equation}
\begin{equation}
x = \mathcal{L}^T x_{\text{low}} + \mathcal{H}^T x_{\text{high}},
\end{equation}

where
\[
\mathcal{L} = \{ l_{n-2}, \ldots, l_{n-2k} \}^T \quad \text{and} \quad \mathcal{H} = \{ h_{n-2}, \ldots, h_{n-2k} \}^T,
\]
the sequences
\[
l_j = \{ l_i \}_{i = j - 2k, \ldots, j + n - 2k} \quad \text{and} \quad h_j = \{ h_i \}_{i = j - 2k, \ldots, j + n - 2k}
\]
represent the low-pass and high-pass filters of an orthogonal wavelet respectively, \(n\) signifies the length of the input vector, \(k\) indicates the length of the frequency component.

\textbf{2D DWT/IDWT.} The DWT applies a 1D DWT to each row and column when getting a 2D data X, i.e.,
\begin{align}
X_{ll} &= \mathcal{L} X \mathcal{L}^T, \quad X_{lh} = \mathcal{H} X \mathcal{L}^T, \\
X_{hl} &= \mathcal{L} X \mathcal{H}^T, \quad X_{hh} = \mathcal{H} X \mathcal{H}^T,
\end{align}
the corresponding 2D (IDWT) is carried out using
\begin{equation}
X = \mathcal{L}^T X_{ll} \mathcal{L} + \mathcal{H}^T X_{lh} \mathcal{L} + \mathcal{L}^T X_{hl} \mathcal{H} + \mathcal{H}^T X_{hh} \mathcal{H},
\end{equation}
where $X_{lh}$, $X_{hl}$, and $X_{hh}$ are the three high-frequency components that preserve the horizontal, vertical, and diagonal details of $X$ respectively when $X_{ll}$ is the low-frequency component of the input $X$, representing the main information, including the fundamental structure of the object.

\textbf{Wavelet Pooling.} 
We choose the Daubechies wavelet transform, namely the Haar wavelet, for Wavelet Pooling. The Haar wavelet transform primarily employs one low-pass filter and n high-pass filters to compute the discrete wavelet transform (DWT) and its inverse (IDWT). In Daubechies-based Wavelet Pooling, a two-dimensional DWT is first applied to decompose the incoming event frame or the feature map generated from the upper-layer CNN into high-frequency detail subbands \( X_{lh} \), \( X_{hl} \), and \( X_{hh} \) as well as a low-frequency approximation subband \( X_{ll} \) in the wavelet domain. A straightforward denoising strategy is then employed, discarding the extracted high-frequency detail subbands \( X_{lh} \), \( X_{hl} \), and \( X_{hh} \), which tend to carry noisy artifacts, and applying the IDWT only on the low-frequency sub-band \( X_{ll} \). This approach reduces the resolution by half, achieving a downsampling effect. At the same time, it captures low-frequency information from the original image or feature map while filtering out high-frequency information without aliasing. Therefore, this method does not increase the number of network parameters significantly.

Based on the Wavelet Pooling using Daubechies, WaveCNet \cite{WaveCNet_2021} utilizes WaveletPooling to replace all downsampling block in network. As formulated in Eq. \ref{eq:WaveCNet}, WaveCNet substitutes WaveletPooling for both max pooling and average pooling, and employs a stride-1 convolution in conjunction with WaveletPooling to replace the stride-2 3x3 convolution.
\begin{align}
\text{MaxPools}_{s=2} \; \rightarrow \; \text{WaveletPool}_{X_{ll}}, \nonumber  \\
\text{Convs}_{s=2} \; \rightarrow \; \text{WaveletPool}_{X_{ll}} \circ \text{Convs}_{s=1}, \nonumber \\
\text{AvgPools}_{s=2} \; \rightarrow \; \text{WaveletPool}_{X_{ll}},
\label{eq:WaveCNet}
\end{align}
where Pool and Conv denote pooling and convolution respectively and $\circ$ signifies followed by. As demonstrated by the experiments in \cite{Vasconcelos_2021_ICCV}, the sequence of feature maps passing through convolution layers and low-pass filters significantly impacts feature extraction. For standard RGB images, applying a low-pass filter before convolution often yields better results. However, for event images reconstructed from event streams, it appears more effective to apply the low-pass filter prior to convolution. This difference may be attributed to the comparatively lower information content in event images than in RGB images. Furthermore, the work in \cite{pmlr-v222-ning24a} showed in the Fig. \ref{fig:WT-Repconv}(b), in denoising networks, maintaining the same sequence of convolution and low-pass filtering across different branches in ResNet blocks yields superior results compared to using different sequences. Most denoising networks adhere to these two principles. We have proposed an improved configuration, as shown in Fig. \ref{fig:WT-Repconv}(c), which our experiments have demonstrated to be more suitable for processing images reconstructed from event streams in Table \ref{tb:fusion_wavelet}.

\subsection{Efficient Hybrid Encoder}
\label{subsec:encoder}
\begin{figure}[btp!]
    \centering
    \includegraphics[width=.7\linewidth]{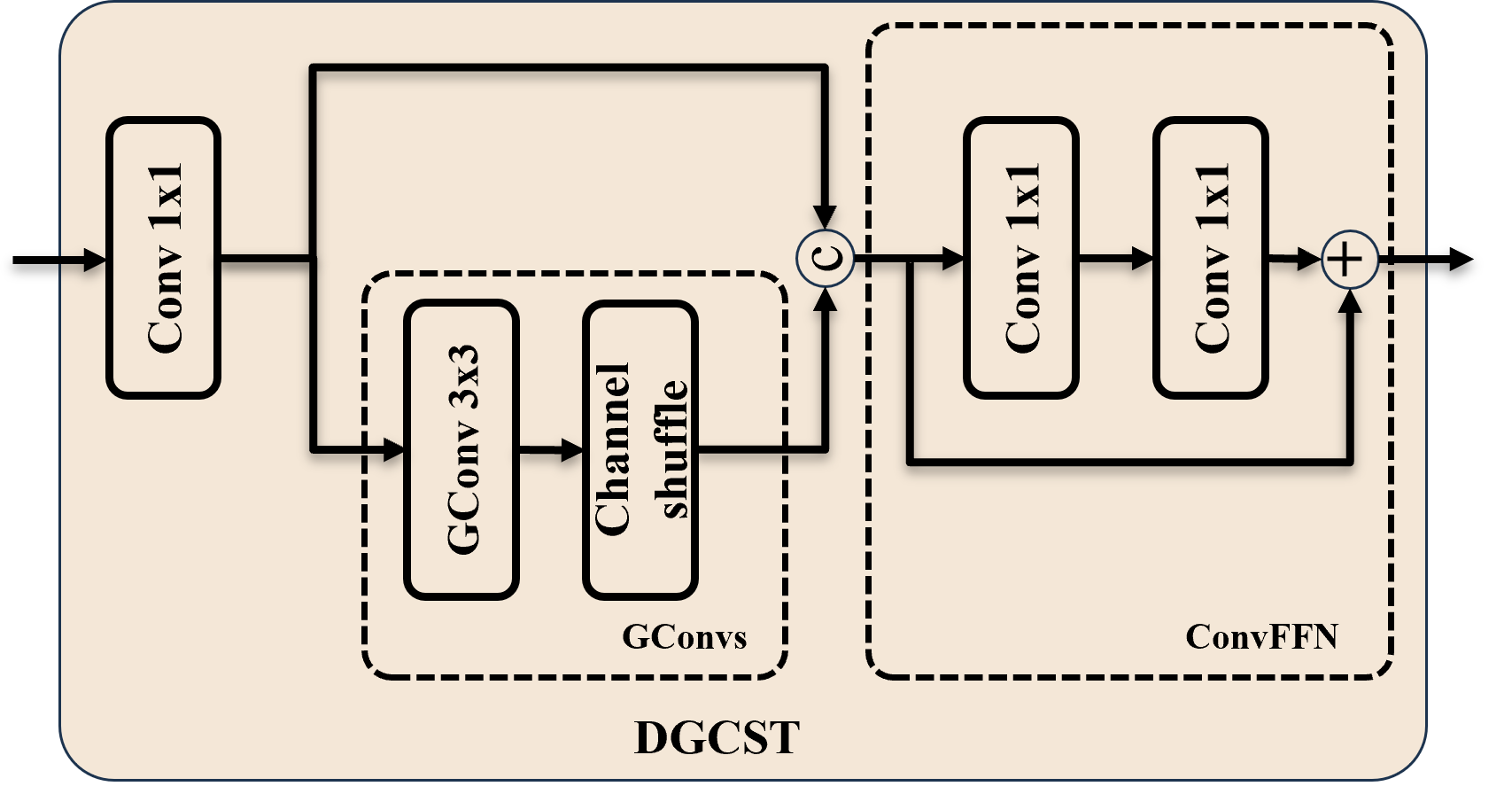}
    \caption{The DRCB block in fusion. The DRCB block integrates depthwise grouped convolutions, channel shuffle, and lightweight feedforward networks for efficient feature fusion.}
    \label{fig:DRCB}
    \vspace{-0.5cm}
\end{figure}

The last three stage features extracted by the backbone is fed into the transformer network, which is composed with encoder and decoder. The encoder is used for feature enhancement and the decoder is used for output the bounding boxes and classifications. In the paper, the basic transformer network is inspired by \cite{Zhao_2024_CVPR}. 

As discussed in the Hybrid Encoder module consists of two modules, including attention-based Intra-scale Feature Interaction (AIFI) and the CNN-based Cross-scale Feature Fusion (CCFF). Three scales of feature maps \( S_3 \), \( S_4 \) and \( S_5 \) can be obtained through backbone. To reduce computational costs, AIFI focuses on the higher-level feature \( S_5 \), which contains richer semantic concepts. This approach also avoids the risk of redundancy and confusion caused by intra-scale interactions with lower-level features. Experiments in \cite{Zhao_2024_CVPR} also provide strong evidence to support this. The CCFF module introduces multiple fusion modules in the fusion pathway that integrate features from adjacent scales into new features, followed by repeated convolution operations for feature enhancement, thus achieving cross-scale fusion. However, we notice that while the repeated convolution operations can enhance features, they also significantly increase the model's parameter count, thereby impacting the final inference speed of the model. 

To address this problem, we proposed the Dynamic Reorganization Conv Block (DRCB), which substituts the repetitive $1 \times 1$ and $3 \times 3$ convolutions in the hybrid encoder. As shown in Fig. \ref{fig:DRCB}, DRCB draws inspiration from the concepts of ShuffleNet V2 \cite{ShuffleNetV2} and employs a $3:1$ strategy for channel separation of feature maps. To reduce the network's parameters and computational requirements, DRCB incorporates group convolution, which also helps prevent overfitting and maintain the network's robustness and generalization capabilities. Additionally, DRCB employs channel shuffle techniques to facilitate the exchange of feature information between groups, ensuring the diversity and richness of the features. Finally, convolutional operations are utilized in place of fully connected layers to achieve similar effects. In all, the proposed DRCB block significantly improves the feature enhancement capability for cross-scale features and further improve the real-time detection performance.

\subsection{Matching Part}
\label{subsec:decoder}

The matching part in WD-DETR consists of three integral components: IoU-aware query selection, Decoder, and Head. By representing the discrepancy between the localization $P$ and the classification $C$ as feature uncertainty $U$ , as shown in Eq. \ref{eq:IOU}, IoU-aware query selection incorporates $U$ into the loss function. This integration combines the bounding box loss $L_{box}$ and classification loss $L_{cls}$ , as shown in Eq. \ref{eq:IOU-L}, to accelerate convergence and improve performance.
\begin{equation}
U(\hat{X}) = \|P(\hat{X}) - C(\hat{X})\|, \quad \hat{X} \in \mathbb{R}^D,
\label{eq:IOU}
\end{equation}
\begin{equation}
L(\hat{X}, \hat{Y}, Y) = L_{\text{box}}(\hat{b}, b) + L_{\text{cls}}(U(\hat{X}), \hat{c}, c),
\label{eq:IOU-L}
\end{equation}
where $\hat{Y}$ represents the predicted values, $\hat{c}$ and $\hat{b}$ represent the predicted class and bounding box, respectively. $Y$ is the ground truth, and $X$ represents the encoder features. 

The decoder comprises six decoding layers, each incorporating a self-attention mechanism and a cross-attention mechanism. The self-attention feature extraction module consists of a self-attention mechanism and a feedforward network, structured similarly to the FFN in the encoder. Unlike the encoder, however, the cross-attention feature extraction module derives its keys and queries from output position encodings, while the values are extracted from the global features produced by the encoder's final layer.

Finally, in the detection head of WD-DETR, the output of the decoder is mapped to the class space through a fully connected layer, generating a probability distribution for the class of each target. The bounding box regression layer then outputs the offsets for the predicted bounding boxes relative to the query positions. The Hungarian algorithm is subsequently applied to match the decoder’s outputs with the ground truth, optimizing the predicted bounding boxes by aligning them with their corresponding true targets.

\begin{figure*}[t!]
    \centering
    \includegraphics[width=1.0\linewidth]{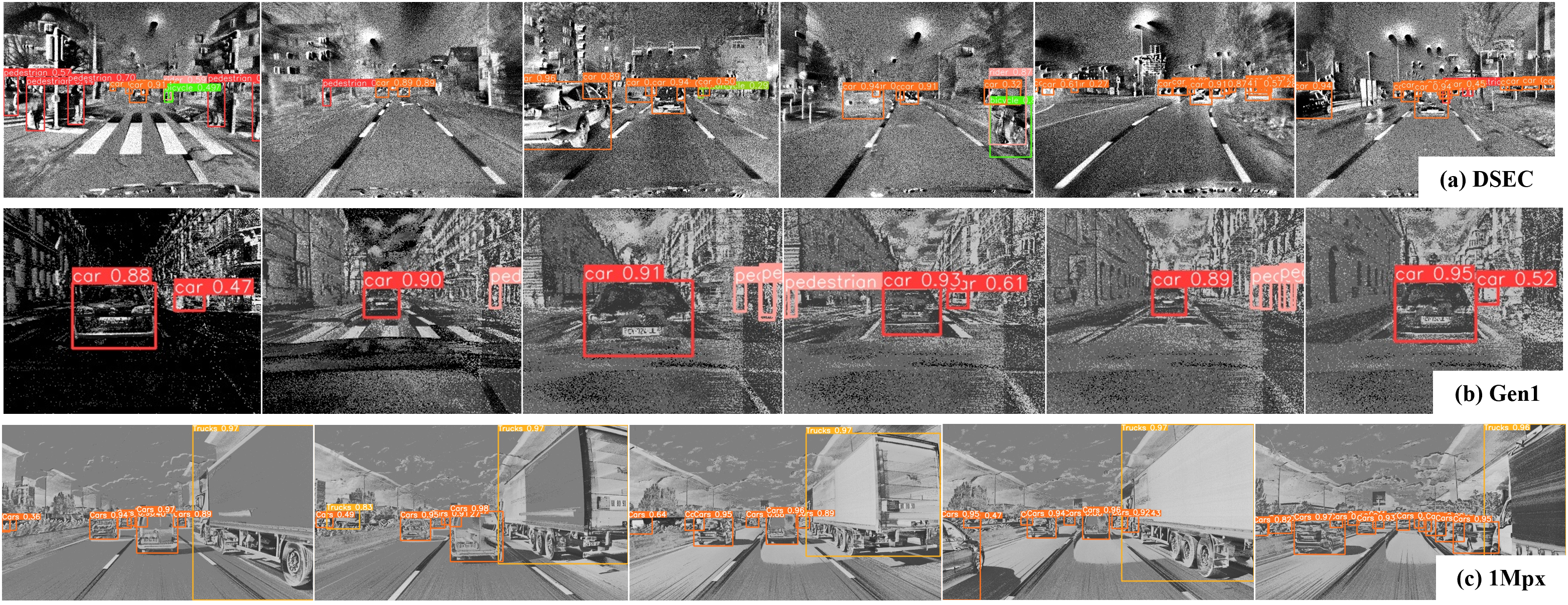}
    \caption{Visualization of object detection results on datasets by WD-DETR, in which event stream is densely represented again, adapted to various bandwidths. The WD-DETR not only could accurately detect small objects in the scene, but it can also effectively identify occluded objects in dense scenes. }
    \label{fig:vis_res}
    \vspace{-0.5cm}
\end{figure*}

\section{EXPERIMENTS}
\label{sec:experiments}

In this section, we first introduce the experimental setup (\ref{subsec:exp_setup}) and the experimental results (\ref{subsec:exp_res}) on the datasets. Then, ablation studies (\ref{subsec:exp_ablation}) are conducted to evaluate our model. Finally, the onboard flight experiment (\ref{subsec:exp_flight}) is conducted to show that our model could keep real-time object detection on the airborne computer. 

\subsection{Experimental Setup}
\label{subsec:exp_setup}

\textbf{Implementation Details.}
We perform all our experiments on a desktop with an Intel 14 vCPU Intel(R) Xeon(R) Platinum 8362 CPU @ 2.80GHz and RTK 3090 GPU. 
All the layers are randomly initialized. The total network is trained with optimizer AdamW. The learning rate is 0.0001. The loss function here is GIOU loss, L1 loss and classification loss.

To construct event representations,  the event stream is represented according to the timestamps of the annotated bounding box and store the event representation as images. And the corresponding bounding box format is re-modified to the YOLO format. In the event representation, the decay factor is $1 \times 10^{-6}$.

\textbf{Datasets.}
In this paper, WD-DETR is evaluated on three event-based object detection datasets, DSEC \cite{DSEC-Dataset}, Gen1 \cite{Gen1-Dataset} and 1Mpx \cite{1mpx_dataset}, which are are widely used for experiments on event cameras.

The DSEC dataset \cite{DSEC-Dataset} is a stereo event camera dataset for driving scenarios, which is the first high-resolution large-scale stereoscopic dataset with an event camera. The resolution of the dataset is $640 \times 480$, containing 53 sequences that drive under a variety of lighting conditions.
The Gen1 Automotive Detection dataset \cite{Gen1-Dataset} is composed of more than 39 hours of automotive recordings acquired with a 304 $\times$ 240 ATIS sensor. The bounding boxes are manually annotated for pedestrians and cars at a frequency between 1 and 4 Hz.
The 1Mpx dataset \cite{1mpx_dataset} contains over 14 hours of scenes captured by event camera. The resolution of the dataset is $1280 \times 720$, and it contains 25 million high-frequency annotated bounding boxes of cars, pedestrians, and two-wheelers. Since the ground truth is obtained automatically by applying an RGB-based detector in the stereo recording setting, there are some geometric errors such as misalignment and semantic errors caused by wrong detection of the detector with the ground truth. In this paper, we use the dataset preprocessed in \cite{DMANnet_2023}, which remove mosaic events and incorrect labels. 

\begin{table*}[htbp!]
\centering
\caption{Comparison studies on Gen1 dataset and 1Mpx dataset. Best results in bold. Runtime is measured in milliseconds for a batch size of 1. A T4 GPU is used for WD-DETR to compare against indicated timings in prior work. } 
\begin{tabular}{ccccccc}
\hline
                 & \multirow{2}{*}{Event Representation} & \multirow{2}{*}{Network Description} & \multicolumn{2}{c}{mAP} & \multirow{2}{*}{Inference Times (ms)} & \multirow{2}{*}{Params (M)} \\
                 &                                       &                                      & Gen1            & 1Mpx  &                                       &                             \\ \hline
YOLOv3 Events    & 2D Histogram                          & CNN                                  & 31.2           & -     & 22.3      & \textgreater{}60      \\
RVT-B            & 2D Histogram                          & Transformer+RNN                      & 47.2           & 47.4 & 10.2      & 18.5                  \\
S4D-ViT-B        & State Space Model                     & Transformer+SSM                      & 46.2           & 46.8 & \textbf{9.4}       & \textbf{16.5}                  \\
GET-T            & Event Histograms                      & Transformer+RNN                      & 47.9           & 48.4 & 16.8      & 21.9                  \\
ERGO-12          & ERGO-12                               & Transformer                          & 50.4           & 40.6 & 69.9      & 59.6                  \\
\textbf{WD-DETR} & Time Decay Event Gray                 & Transformer+CNN                      & \textbf{52.8}  & \textbf{65.5}      & 32.3      & 34.9                 \\ \hline
\end{tabular}
\label{tb:com_gen1}
\end{table*}

\begin{table*}[htbp!]
\centering
\caption{Ablation studies on DSEC dataset. WaveletPooling demonstrates the most significant improvement in mAP, while DRCB achieves the greatest reduction in inference time.} 
\begin{tabular}{ccccccccc}
\hline
WaveletPooling            & RepConv                   & DRCB                     & mAP$_{50}$ & mAP$_{50-95}$ & Layers & Params (M) & GFLOPs & Inference Time (ms) \\ \hline
×                         & ×                         & ×                         & 36       & 22.2         & 479    & 41.97      & 129.6  &  31.7                   \\
×                         & \checkmark                & \checkmark                & 34.2     & 20.2        & 445    & 34.91      & 90.4  & 27.0                    \\
\checkmark                & ×                         & \checkmark                & 34.5      & 20.1         & 442    & 34.92      & 107.6  &  31.7                   \\
\checkmark                & \checkmark                & ×                         & 36.4     & 22.4        & 435    & 41.97      & 142.8   &  40.8                   \\
\checkmark                & \checkmark                & \checkmark                & 38.6     & 23.5        & 443    & 34.92      & 107.6  &  32.3                \\ \hline
\end{tabular}
\label{tb:ablation}
\end{table*}

\begin{table}[tbp!]
\caption{Comparison studies on DSEC dataset. The proposed WD-DETR method achieves the highest success rate with only events as input, comparable to the performance of the CAFR method, which fuses events and frames. } 
\begin{tabular}{cccc}
\hline
\multirow{2}{*}{Methods}     & \multirow{2}{*}{Input} & Network                                                                        & \multirow{2}{*}{mAP} \\
                             &                        & Description                                                                    &                      \\ \hline
\multirow{3}{*}{CAFR}        & Events                 & \multirow{3}{*}{ResNet50+CAFR}                                                 & 12                 \\
                             & Frames                 &                                                                                & 25                 \\
                             & Events + Frames        &                                                                                & \textbf{38}                 \\ \hline
EFNet                        & Events+Frames          & \begin{tabular}[c]{@{}c@{}}UNet-like Backbone+\\ Attention module\end{tabular} & 30                 \\ \hline
\multirow{2}{*}{RetinaNet50} & Events                 & \multirow{2}{*}{\begin{tabular}[c]{@{}c@{}}ResNet50+\\ RetinaNet\end{tabular}} & 12                 \\
                             & Events + Frames        &                                                                                & 26                 \\ \hline
RENet                        & Events + Frames        & \begin{tabular}[c]{@{}c@{}}ResNet+ BDE Fusion\\ Decoder\end{tabular}           & 29.4                \\ \hline
\textbf{WD-DETR}                      & Events                 & \begin{tabular}[c]{@{}c@{}}WP-Repconv Backbone\\  + DETR\end{tabular}         & \textbf{38.4}                \\ \hline
\end{tabular}
\label{tb:com_dsec}
\vspace{-0.2 cm}
\end{table}

\subsection{Experimental Results}
\label{subsec:exp_res}
We compare WE-DETR with other state-of-the-art methods on DSEC, Gen1 and 1Mpx dataset, and report the results in Table \ref{tb:com_dsec} and Table \ref{tb:com_gen1}. CAFR \cite{CAFR_2024}, EFNet \cite{EFNet_2022}, RetinaNet50 \cite{Retina-Net_2022} and RENet \cite{RENet} and included in the comparison on DSEC dataset. YOLOv3 Events \cite{YOLOV3_Events_2019}, RVT-B \cite{RecurrentTransformerNet}, S4D-ViT-B\cite{SSM_2024}, GET-T \cite{GET-Net}, ERGO12 \cite{GWD-Net} are included in the comparison on Gen1 datasaet and 1Mpx dataset.

When only event data are used, the accuracy of our method on DSEC data set is superior to other state-of-the-art methods, and the accuracy is close to that of CAFR method using event and image fusion. The results show that dense event representations with rich texture could replace the role of color images in object detection to a certain extent.

WE-DETR achieves a new state-of-the-art performance of $52.8$ mAP on the Gen1 dataset and $65.5$ mAP on the 1 Mpx dataset. ERGO-12 claims comparable results on both datasets, albeit at the cost of using a much larger network and increased inference time. The GET-T model also reports favorable results but achieves a $4.9$ lower mAP on the Gen1 and $15.7$ lower mAP on the 1Mpx dataset compared with our model. For inference time, although our method is not the fastest, it achieves an inference time that is nearly half of that of ERGO-12 while maintaining a higher accuracy than ERGO-12. As a result, WE-DETR performed significantly better on the 1Mpx dataset than other algorithms, and part of the reason for this may be that the 1Mpx dataset we used was preprocessed by \cite{DMANnet_2023}, which removed mosaic events and incorrect labels. 
Finally, some object detection results predicted by WD-DETR are presented in Fig. \ref{fig:vis_res}
, demonstrating that WD-DETR is also capable of detecting small targets and occluded targets in dense scenes.

\subsection{Ablation Studies}
\label{subsec:exp_ablation}
In this section, we examine the three main contributors, namely, WaveletPooling block, RepConv block, and DRCB, to the final performance of the proposed model on DSEC dataset. The results of ablation studies is shown in Table \ref{tb:ablation}. Without WaveletPooling block, the mAP$_{50}$ decreases the most, which indicates that WaveletPooling block to some extent reduces the influence of noise on feature extraction. Without DRCB, the model size increases by nearly 7M, and the inference time increases by 8ms, which shows that this module indeed reduces the network size and speeds up the network operation. In addition, modifying WaveletPooling and RepConv in the baseline without increasing the network parameters and inference time improves the network accuracy.

In addition, we also compare the network architecture of the backbone for wavelet pooling fusion method, shown in Fig. \ref{fig:WT-Repconv}. The result is shown in Table \ref{tb:fusion_wavelet}. Under the same set of training parameters, although the number of network layers and the size of the parameters are consistent, our fusion method can achieve a higher mAP on the DSEC dataset after the same number of training epochs and batch size.

\begin{table}[tbp!]
\centering
\caption{Comparison of different fusion methods of wavelet pooling in the backbone.}
\begin{tabular}{ccccc}
\hline
               & mAP$_{50}$ & mAP$_{50-95}$ & Layers & GFLOPs \\ \hline
Our block      & \textbf{0.386} & \textbf{0.235}    & 443    & 107.6  \\
Previous Block & 0.376 & 0.224    & 443    & 130.5  \\ \hline
\end{tabular}
\label{tb:fusion_wavelet}
\end{table}

\subsection{Robotic Onboard Computer Experiment}
\label{subsec:exp_flight}

\begin{figure*}[htbp!]
    \centering
    \includegraphics[width=1.0\linewidth]{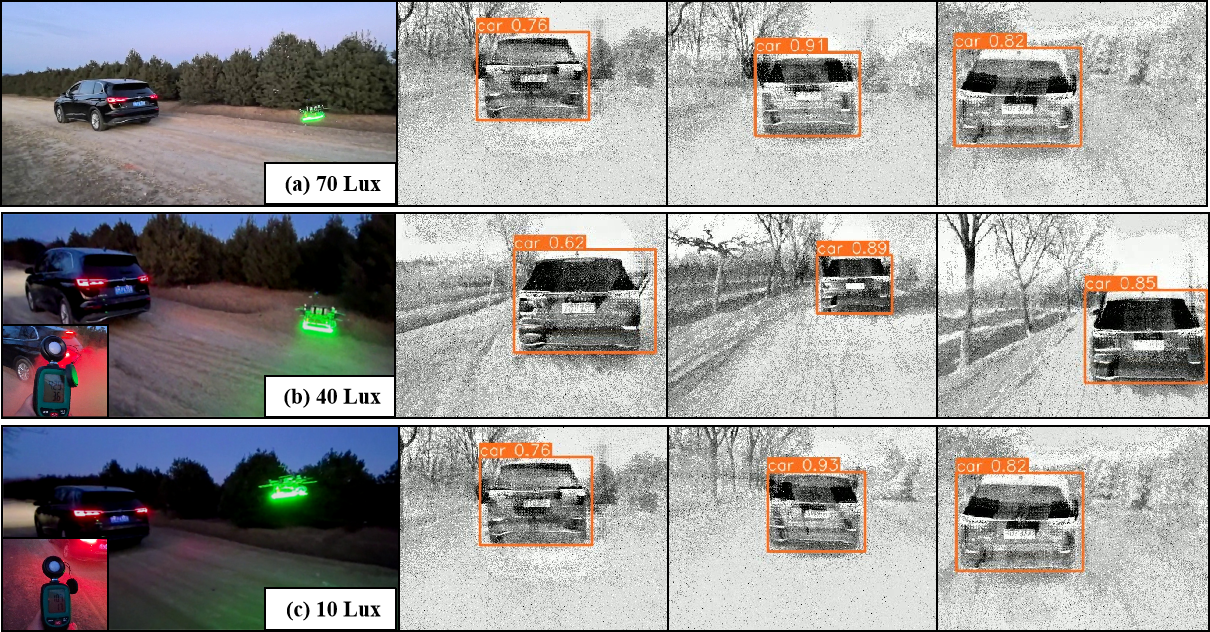}
    \caption{Outdoor flight experiment results, including the thrid view of experiment environment and the detection results of WD-DETR. The weights trained by DSEC dataset is used for object detection. The detection threshold is set to 0.5. The flight experiments of object detection are carried out under the light intensity of 70-80, 40-50 and 10-20 Lux.}
    \label{fig:outdoor_exp2}
    \vspace{-0.5cm}
\end{figure*}

The WD-DETR model is converted to TensorRT in FP16 precision to enhance inference performance on airborne computer, NVIDIA Jetson Orin NX. 

NVIDIA Jetson Orin NX supports different power consumption modes, which correspond to different performance configurations to adapt to different application scenarios and power budgets. At a power consumption of 25W, 8 CPU cores and 4 GPU TPCs were enabled, with a maximum CPU frequency of 1497.6 MHz and a maximum GPU frequency of 408 MHz. In the unlimited power mode (MAXN), 8 CPU cores and 4 GPU TPCs were enabled, with a maximum CPU frequency of 1984 MHz and a maximum GPU frequency of 918 MHz. The input of the network is the event representation with a size of $640 \times 640$. The batch size is set to 1. 
The FPS and memory usage results under different modes are shown in Table \ref{tb:onboard}.

At a mode of 25W on the onboard computer, the FPS of the proposed method is $16.52$. When the mode is set to max, the FPS can reach $35.19$. Therefore, the WD-DETR framework can be performed in real-time on the robot's onboard computer for the event camera, enabling efficient object detection tasks.

\begin{figure}[tbp!]
    \centering
    \includegraphics[width=0.8\linewidth]{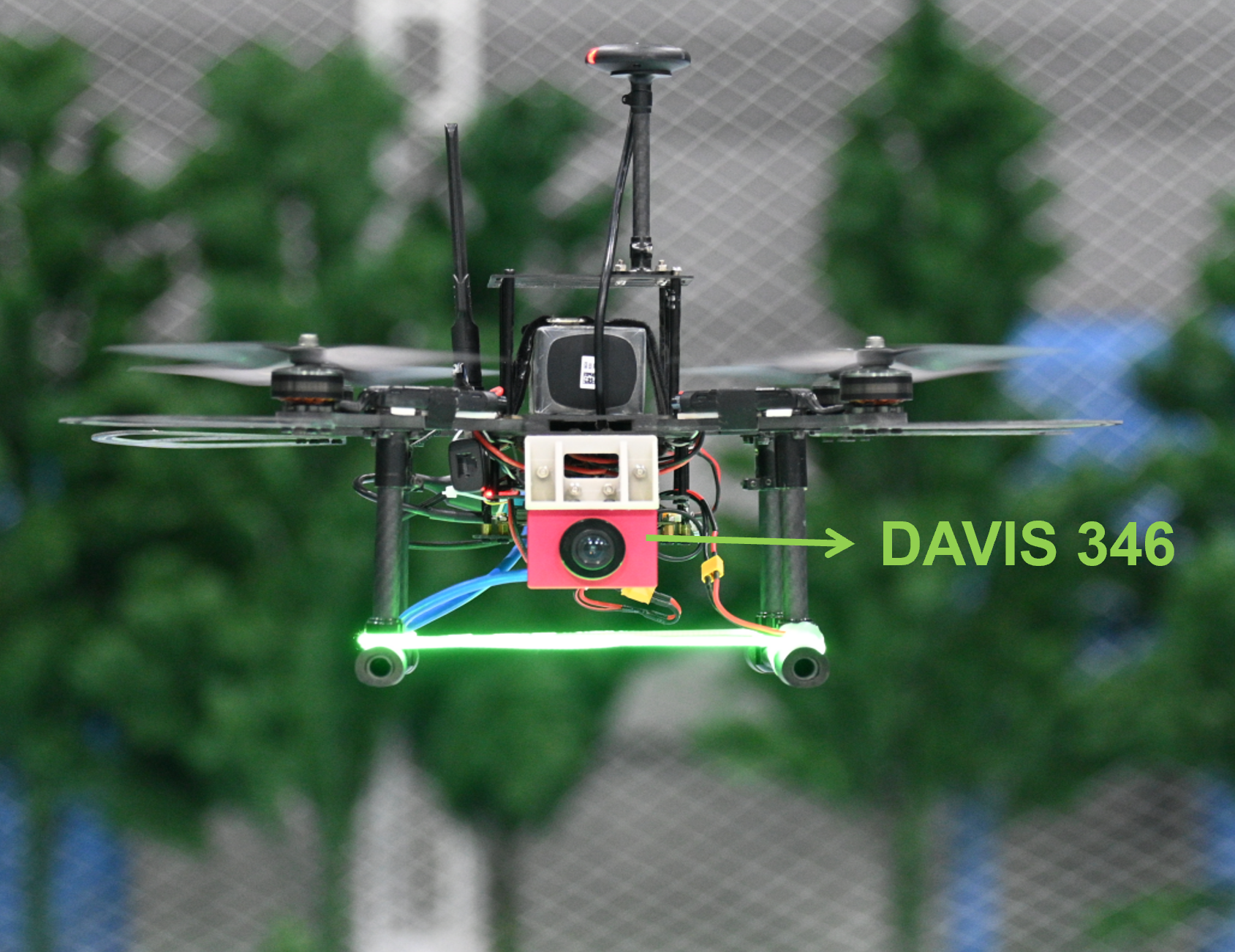}
    \caption{The UAV utilized in the outdoor flight experiment is equipped with a forward-looking event camera, the DAVIS 346, and a NVIDIA Jetson Orin NX for real-time object detection.}
    \label{fig:outdoor_exp1}
    \vspace{-0.2cm}
\end{figure}

\begin{table}[htbp!]
\centering
\caption{Performance of different modes on the onboard computer NVIDIA Jetson Orin NX utilizing TensorRT FP 16.}
\begin{tabular}{ccccc}
\hline
Modes & FPS   & Mem        & GPU           & CPU   \\ \hline
25W   & 16.52 & 2.6G/15.3G & 400MHz 98.7\% & 7.6\% \\
MAXN  & 35.19 & 2.6G/15.3G & 910MHz 98.4\% & 8.4\% \\ \hline
\end{tabular}
\label{tb:onboard}
\vspace{-0.2cm}
\end{table}

\begin{table}[htbp!]
\centering
\caption{specific performance of event camera DAVIS346}
\begin{tabular}{ll}
\hline
Resolution                         & 346 $\times$ 260   \\ 
Weight                             & 100 g        \\
Size (cm)                          & 4 $\times$ 6 $\times$ 2.5   \\
Dynamic Range (dB)                 & 120        \\
Latency (us)                       & 20          \\
Stationary noise (ev/pix/s) at 25C & 0.12        \\
FOV of 1/3 inch lens (Degree)      & 86$\times$72$\times$51  \\ \hline
\end{tabular}
\label{tb:DAVIS346}
\vspace{-0.2cm}
\end{table}

To evaluate the proposed method on airborne computer, the platform for flight experiment is shown as Fig. \ref{fig:outdoor_exp1}, which is composed with a flight controller, event camera DAVIS 346 and the airborne computer NVIDIA Jetson Orin NX. The DAIVS346 is an event camera from Inivation Company that outputs both event data and RGB images. Table \ref{tb:DAVIS346} lists the specific performance indicators. The NVIDIA Jetson Orin NX is equipped with an NVIDIA Ampere architecture-based GPU, featuring 1024 CUDA cores and 32 Tensor cores. It is powered by an Arm Cortex-A78AE CPU and comes with 8 CPU cores. 

The outdoor flight experiments were conducted under light intensities of 70-80 Lux, 40-50 Lux, and 10-20 Lux. The event camera captured the event stream and detected objects in real time using the onboard computer. The third perspective of the experiments, along with the detection results from the event image representation, is illustrated in Fig. \ref{fig:outdoor_exp2}. The detection threshold for WE-DETR was set at 0.5.
At lower light intensities, the event stream contains a higher level of noise. This phenomenon occurs because the event camera records logarithmic intensity changes rather than absolute brightness values. Under low-light conditions, noise events are triggered more frequently due to greater variance in logarithmic intensity variations. Conversely, under high-brightness conditions, there is a relatively lower occurrence of noise events.
The event representation method and object detection approach utilized by WD-DETR, as proposed in this paper, effectively represent events as images and detect objects even under low light intensity conditions of approximately 10 Lux.

\section{CONCLUSIONS}
\label{sec:conclusions}
In this work, we present WD-DETR, a novel object detection transformer network designed for event cameras, enhanced with wavelet-based denoising. The proposed architecture is composed of four key components: a dense event representation module, a wavelet-enhanced backbone, an efficient hybrid encoder, and a matching module. 
First, the dense event representation module transforms sparse event data into a dense tensor in real-time. Next, the wavelet-enhanced backbone integrates a de-noising algorithm directly into the feature extraction process, effectively filtering noise while capturing event features. Subsequently, the efficient hybrid encoder, leveraging the DRCB model, further refines the extracted features. Finally, the matching module predicts the detected objects.
Experimental results on benchmark datasets demonstrate that WD-DETR outperforms state-of-the-art methods in object detection using event cameras. Additionally, the model is optimized for deployment by converting it to TensorRT FP16 and running it on a standard onboard robotic computer, achieving an impressive 35 FPS, which is highly suitable for robotic systems engaged in real-time object detection. 
The WD-DETR demonstrated the capability to effectively detect objects in real-time during outdoor flight experiments conducted under low light conditions.

\bibliographystyle{IEEEtran}
\bibliography{ref}

\vfill

\end{document}